\documentclass[11pt]{article}

\usepackage[final]{acl}
\usepackage{censor}
\usepackage{times}
\usepackage{latexsym}

\usepackage{makecell}
\usepackage{tabularx,array}
\usepackage{float}

\usepackage[T1]{fontenc}

\usepackage[utf8]{inputenc}

\usepackage{enumitem}
\usepackage{ragged2e}
\usepackage{microtype}

\usepackage{inconsolata}

\usepackage{graphicx}

\title{Less is More: Quality-Aware Training Data Selection \\ for Scientific Summarization}

\author{
\textbf{Maria Nefeli Paraskevopoulou, Tatiana Passali and Grigorios Tsoumakas} \\
School of Informatics \\
Aristotle University of Thessaloniki \\
Greece \\
\texttt{\{mparaskf,scpassali,greg\}@csd.auth.gr}
}

\begin{document}
\maketitle
\begin{abstract}
Scientific long-document summarization datasets commonly treat author-written abstracts as gold reference summaries, although their quality and alignment with the source article vary. At the same time, publicly available scientific summarization datasets remain limited in scale and structure for modern long-context models. In this work, we address both challenges by a) constructing and releasing 
one of the largest biomedical and life science datasets for long-document summarization, containing 1.88 million PMC articles, and b) analyzing the reference quality of author-written abstracts with source-grounded and model-based metrics. We show that author-written abstracts vary in their alignment with the full article and that these quality signals can guide training-data selection. Training on selected high-quality subsets outperforms random sampling at matched training sizes and can match or exceed larger random subsets on factuality-oriented metrics. Our findings suggest that reference quality is an important factor in scientific summarization and that quality-aware data selection can improve training efficiency.
\end{abstract}

\section{Introduction}
Over 1.5 million biomedical and life science articles are indexed in PubMed annually, making it increasingly difficult for researchers and practitioners to keep up with the scientific literature~\citep{gonzalez2024landscape}. This has motivated long-document abstractive summarization systems designed to generate concise summaries of full papers~\citep{cohan-etal-2018-discourse}. However, the effectiveness and reliability of such systems depend critically on the quality of the data used for training and evaluation.

Most scientific summarization benchmarks pair each article with its author-written abstract and treat the abstract as the reference summary \cite{gupta-etal-2021-sumpubmed}. Yet, prior work in biomedical and healthcare research has shown that abstracts can diverge from their corresponding full texts through omissions, inconsistencies, selective reporting, or overly strong interpretations of the findings \citep{pitkin1999accuracy}. These mismatches matter because abstracts are used both as supervision during model training and as reference targets during evaluation. If reference summaries vary substantially in quality, then models trained on them may learn to generate unreliable summaries, and evaluation against them may reward behavior that does not fully reflect source-grounded factuality~\citep{zhang-etal-2024-benchmarking}.

Beyond concerns about reference quality, existing public benchmarks for biomedical and life science article summarization remain limited. The PubMed benchmark \cite{cohan-etal-2018-discourse} remains one of the most widely used resources for the task, with several later resources reusing it~\citep{zaheer2020bigbird, huang-etal-2021-efficient, pang2023long} or adapting it for related settings such as aspect-based scientific document summarization \citep{soleimani-etal-2022-zero}. However, it was introduced before the emergence of long-context Large Language Models (LLMs), and does not preserve hierarchical document structure in a format optimized for such models~\citep{sclar2024quantifying,he2024does,li2022markuplm}. Together, these limitations raise concerns about both the reliability of supervision signals and the adequacy of current benchmarks for modern long-document summarization systems.

To address these limitations, we build and release PMC-Large\footnote{\href{https://huggingface.co/datasets/nefelipar/pmc-large}{https://huggingface.co/datasets/nefelipar/pmc-large}}, a large-scale benchmark from PubMed Central (PMC) Open Access research articles,  containing 1.88 million full papers. Each article is paired with its abstract, and the body is encoded in a Markdown-style format that preserves section hierarchy, filters non-narrative content, and normalizes non-textual elements with placeholders. Using a subset of this dataset, we investigate whether the quality of author-written abstracts varies across the corpus.  We define quality as the degree to which an author-written abstract is supported by the corresponding article body, as measured by source-grounded evaluation metrics, including AlignScore~\citep{zha-etal-2023-alignscore}, FineSurE~\citep{song-etal-2024-finesure}, G-Eval~\citep{liu-etal-2023-g}, and SummaC~\cite{laban-etal-2022-summac}. Finally, we test whether selecting higher-quality examples for fine-tuning leads to better models than training on random samples. Our experiments reveal that quality-selected subsets outperform random subsets at matched training sizes, and smaller quality-selected subsets can match or exceed larger random ones on factuality-oriented metrics.

Our main contributions are as follows:
\begin{itemize}[noitemsep, topsep=0pt]
\item We release PMC-Large, a large-scale long-document summarization dataset, 
containing 1.88M PMC article-abstract pairs, and preserving article structure for modern long-context LLMs.
\item We analyze the reference quality of author-written abstracts on a 10,000-article subset using multiple evaluation metrics.
\item We show that author-written abstracts vary in source-grounded reference quality, and that quality-based data selection outperforms random subsets at matched training sizes and can match or exceed larger random subsets.

\end{itemize}


\section{Related Work}
\label{sec:related_work}

\subsection{Scientific summarization}
Scientific summarization aims to produce concise summaries of research articles while preserving the main scientific findings. Biomedical summarization presents additional domain-specific challenges such as specialized terminology, abbreviations, numerical results, methodological details, and a high requirement for factual consistency \citep{mushtaq2026biomedical}. 

The length of scientific articles introduces additional challenges because relevant evidence may be distributed across multiple sections of a paper. Earlier transformer-based models were limited by input length, which often required truncation, chunking or separate content-selection steps \citep{lewis2020bart, zhang2020pegasus}. Existing work addresses these challenges through long-context architectures \citep{beltagy2020longformer, zaheer2020bigbird}, content-selection pipelines such as retrieve-then-summarize \citep{zhang2022summn}, and models that incorporate discourse or section structure \citep{cohan-etal-2018-discourse,10.1109/TASLP.2020.3037401}. 

\subsection{Scientific summarization datasets}
Scientific summarization datasets cover a range of tasks, including lay summarization \citep{guo2021automated, goldsack2022making, goldsack-etal-2023-biolaysumm}, question summarization \citep{abacha2019summarization}, and multi-document evidence synthesis \citep{deyoung2021ms2}. However, these resources are not directly comparable to full-article scientific summarization since they use different input types or different summary targets.

Several article-based PMC summarization datasets define related but distinct task formulations. PMC-SA \citep{gidiotis2019structured} targets section-aware summarization with structured abstract outputs, whereas PubMedCite \citep{luo2023citationsum} augments article inputs with citation-derived context from cited papers. In contrast, our dataset focuses on single-document full-article summarization using only the article body as input and the author abstract as target.

The closest datasets to our work are PubMed and SumPubMed, which both pair PMC article bodies with author-written abstracts. PubMed \citep{cohan-etal-2018-discourse} contains approximately 133K articles and provides article text together with section-level information. SumPubMed \citep{gupta-etal-2021-sumpubmed} contains 34K articles from the BioMed Central collection and follows the same article-to-abstract formulation. Among these, PubMed is the most relevant point of comparison for our work, since both datasets are derived from PubMed Central.

\subsection{Reference quality of abstracts}
Author-written abstracts are commonly treated as gold reference summaries in scientific summarization datasets because they are widely available and provide concise descriptions of the corresponding articles. However, prior work in biomedical and healthcare research shows that abstracts are not always fully consistent with the full text. Studies have documented discrepancies between abstracts and articles, including missing information, inconsistent numerical results, incomplete reporting of harms, and conclusions that are stronger than the evidence supports \citep{pitkin1999accuracy, bernal2008abstracts, li2017scoping, nascimento2021not, kamel2023reporting}.

A related concern is {\em spin}, where the wording or emphasis of an abstract presents findings more favorably than justified by the results. Spin can appear in biomedical abstracts through selective emphasis, causal language, or overconfident interpretation of findings \citep{lazarus2015classification, boutron2018misrepresentation}. Taken together, these studies suggest that the quality of abstracts is variable and that their reliability as reference summaries should be examined instead of assumed.





\subsection{Evaluation of summary quality}

Traditional reference-based metrics, such as ROUGE and BERTScore, compare generated summaries with reference summaries using lexical or semantic similarity, but they do not directly verify whether summary claims are supported by the source document. However, high reference similarity does not necessarily imply factual correctness or complete information overlap \citep{deutsch-roth-2021-understanding,kryscinski2020evaluating}. This limitation is especially important in scientific summarization, where factual errors, unsupported conclusions, or missing evidence can change the interpretation of scientific findings \citep{zhang-etal-2023-famesumm,fang2024understanding}. Recent work has therefore moved toward source-grounded and model-based metrics, such as SummaC \citep{laban-etal-2022-summac}, AlignScore \citep{zha-etal-2023-alignscore}, FineSurE \citep{song-etal-2024-finesure}, and G-Eval \citep{liu-etal-2023-g}, which assess summary quality using the input document as evidence.



\subsection{Training-data selection for summarization}
A related line of work studies how the choice and quality of training examples affect summarization models. \citet{sun2023data} propose a data-selection curriculum for abstractive summarization and show that selecting useful examples can improve training. Other work focuses on noisy or unfaithful supervision. \citet{matsumaru2020improving} filter article-headline pairs using entailment signals to improve the truthfulness of generated headlines, while \citet{nan2021entity} show that filtering training data with factual consistency signals can reduce entity hallucinations in abstractive summarization. \citet{adams2022learning} also show that reference summaries may contain unsupported information and propose revising unfaithful references instead of removing them. These works highlight the importance of supervision quality during training. Our work builds on this motivation in the context of biomedical and life science long-document summarization by analyzing the quality of author-written abstracts at scale and studying whether source-grounded reference-quality signals can guide training-data selection.

\section{Dataset Creation}
\label{sec:dataset}

We construct a large-scale long-document summarization dataset from the PubMed Central Open Access (PMC-OA) corpus, which contains biomedical and life science research articles. The source articles are distributed as Journal Article Tag Suite (JATS)\footnote{\href{https://jats.nlm.nih.gov}{https://jats.nlm.nih.gov}} XML files, which encode article text, abstract, nested sections, citations, equations, figure and table mentions, and publication metadata. Each dataset instance pairs the processed main body of a research article with its author-written abstract. The article body is used as the source document and the abstract as the reference summary.

\subsection{Source and record construction}

We convert PMC-OA XML archives into JSONL records using a preprocessing pipeline. Each record contains the PMC identifier, processed article body, processed abstract, and bibliographic metadata.
We retain only articles with both a non-empty body and a non-empty abstract, since both are required for supervised summarization. We also restrict the corpus to permissively licensed articles (CC BY, CC0) and remove cases where the abstract is detected as duplicated in the article body, avoiding target leakage during supervised fine-tuning.

\subsection{Structure and normalization}

A central goal of the pipeline is to retain scientific article structure in a format suitable for long-context language models. We therefore serialize article bodies in a markdown-style format, preserving section hierarchy and other structural cues directly in the model input. This choice is motivated by evidence that LLMs are sensitive to input formatting and that structured or markup-aware document representations can improve document understanding compared to structure-agnostic plain text \citep{sclar2024quantifying,he2024does,li2022markuplm}. Section and subsection titles are converted into Markdown heading markers, while nested section hierarchy is preserved through heading depth. Paragraphs, lists, boxed text, display quotes, and other textual blocks are serialized in the order in which they appear in the article. This keeps structural information available to the model while adding little token overhead.


We also standardize how scientific elements are represented in the text. Non-textual content such as figures, tables, and supplementary files is removed, while references of citations, figures, tables, appendices, boxed text and
supplementary material are replaced with placeholders (e.g. figure references are replaced with \texttt{[FIG\_REF]}). Formula and code markup are normalized with placeholders, using \texttt{[MATH]} and \texttt{[CODE]}, respectively.
By doing this, the dataset remains focused on the article text while still indicating where references, formulas, and code
occur in the source. Qualitative examples of text normalization are shown in Appendix~\ref{app:preprocessing_examples}, Table~\ref{tab:normalization_examples}. Unlike the dataset of \citet{cohan-etal-2018-discourse}, which applies stronger normalization to scientific notation and symbols, our preprocessing preserves text-encoded symbols, units, numeric ranges, subscript/superscript notation and related scientific notation where possible.

\subsection{Filtering and corpus statistics}


PMC articles contain many sections outside the main scientific narrative, including references, acknowledgements, disclosures, and other publication-related material that is not usually summarized by the abstract. Identifying these sections is not trivial because PMC articles are heterogeneous. Some are marked by dedicated XML tags, others by \texttt{section-type} attributes, and others only by their headings. They also do not appear in a fixed position within the article.

We therefore apply a three-step section filter. First, we remove sections whose XML tags belong to a predefined set of tags that usually indicate extra article material rather than the main content. Second, for general section containers, we inspect the optional JATS \texttt{sec-type} attribute, which provides a functional label for some sections, and remove sections whose type indicates publication-related material such as data-availability, supplementary-material, display-object, or competing-interest sections. Third, when this metadata is incomplete or inconsistent, we standardize section titles and match them against a curated set of regular-expression patterns to identify and remove material that is not summarized by the abstract. The result is a filter that is more robust to the different ways PMC articles encode the same kind of content. Examples of non-narrative sections removed by our filter are shown in Appendix~\ref{app:preprocessing_examples}, Table~\ref{tab:section_filtering_examples}. These include, among others, funding statements, acknowledgements, consent sections, data availability statements, and lists of abbreviations, which are retained in the released PubMed dataset of \citet{cohan-etal-2018-discourse}.

\begin{table}[t]
\centering
\small
\begin{tabular}{lll}
\hline
\textbf{Dataset aspect} & \textbf{PubMed} & \textbf{Ours} \\
\hline
Source & PMC & PMC-OA \\
Target & Author abstract & Author abstract \\
Pairing & article-abstract & article-abstract \\
Size & 133K & 1.88M \\
Avg. doc. length & 3{,}016 & 3{,}918 \\
Avg. sum. length & 203 & 218 \\
Sections & Level-1 & Nested \\
Format & Plain text & Markdown \\
Section filtering & Up to conclusion & Content-aware \\
Figures/tables & Removed & Removed \\
Inline refs & Not specified & Placeholders \\
Math/Citations & Special token & Placeholders \\
Code & Not specified & Placeholders \\
Length filtering & Length-based & Percentile-based \\
Language filter & Not reported & English \\
License filter & Not reported & CC BY/CC0 \\
Abstract leakage & Not reported & Removed \\
\hline
\end{tabular}
\caption{Comparison of dataset-construction choices between the PubMed dataset \citep{cohan-etal-2018-discourse} and ours.}
\label{tab:dataset_construction}
\end{table}

After XML conversion and section filtering, we apply corpus-level filters. We keep only English research articles and retain examples whose body and abstract lengths fall between the 10th and 80th percentiles of their respective token distributions. This choice concentrates the corpus on the central mass of the distribution and removes extremely short articles and very long outliers that could skew training and evaluation. A more detailed analysis of these distributions is provided in Appendix~\ref{app:dataset_statistics},  Figure~\ref{fig:token-count}. Token counts are computed with the \texttt{Qwen2.5-3B} tokenizer, where each placeholder (such as [CIT\_REF]) is counted as a single token. In the final corpus, this corresponds to 3,011-8,750 tokens for the article body and 191-440 tokens for the abstract. 

The initial conversion stage yields 4,407,836 cleaned records, which are reduced to 1,884,517 article-abstract pairs after final filtering. These preprocessing and filtering choices are summarized in comparison with the dataset of \citet{cohan-etal-2018-discourse} in Table~\ref{tab:dataset_construction}. Our dataset preserves richer document structure through nested section representations and Markdown serialization, applies more targeted content-aware filtering, and uses explicit placeholders for mentions
of non-textual elements, formula and code markup. The final corpus is obtained after applying corpus-level filters for language, length, licensing, and cases where the abstract is duplicated in the article body, which could leak the target summary into the input.

\section{Quality Analysis of Abstracts}
\label{sec:quality_analysis}

Before using author-written abstracts as supervision targets, we first evaluate their quality as summaries of the corresponding article body. This experiment asks whether abstracts in PMC can be treated as uniformly reliable reference summaries, or whether their quality varies substantially across examples. We sample 10{,}000 article-abstract pairs from our dataset and treat each abstract as the summary under evaluation, with the corresponding processed article body as the source document.

\subsection{Evaluation metrics}
\label{subsec:metrics}
For each article-abstract pair, we compute the following four source-grounded and
model-based metrics that evaluate both author-written abstracts and generated summaries against the corresponding article body. For the judge models of FineSurE and G-Eval, we use Gemini 2.5 Flash.



\textbf{AlignScore} \cite{zha-etal-2023-alignscore} measures the alignment between a summary and its source document. In our experiments, we use AlignScore to estimate how well author-written abstracts and generated summaries are supported by the corresponding article body.

\textbf{FineSurE} \citep{song-etal-2024-finesure} provides fine-grained factuality evaluation through sentence-level error detection. We use FineSurE to evaluate the faithfulness of author-written abstracts and generated summaries with respect to the article body.

\textbf{SummaC} \cite{laban-etal-2022-summac} is an NLI-based metric for evaluating whether summary statements are supported by the source document. We use the zero-shot variant, SummaC-ZS, with a biomedical NLI backbone, PubMedBERT-MNLI-MedNLI\footnote{\href{https://huggingface.co/pritamdeka/PubMedBERT-MNLI-MedNLI}{https://huggingface.co/pritamdeka/PubMedBERT-MNLI-MedNLI}}. We use this configuration because initial experiments with the default general-domain SummaC settings showed limited suitability for long biomedical and life science articles.

\textbf{G-Eval} \cite{liu-etal-2023-g} uses an LLM judge with prompting to evaluate summary quality according to predefined criteria. We use three criteria, factual consistency, coherence, and relevance, and report their average as the G-Eval score.

Unlike standard summarization settings, the author-written abstract is itself evaluated as candidate summary, rather than treated only as a reference target. Thus, as discussed in Section~\ref{sec:related_work}, we do not use ROUGE or BERTScore because they rely on reference-based similarity and do not directly assess whether summaries are supported by the source document. 


\subsection{Quality analysis}

We examine whether the score distributions exhibit a low-quality tail and whether the metrics agree on which abstracts are high or low quality. This analysis is separate from model training, since it evaluates the reference abstracts directly and motivates subsequent experiments that use quality signals derived by the metrics for training data selection.

\begin{figure}[h]
\centering
\includegraphics[width=\linewidth]{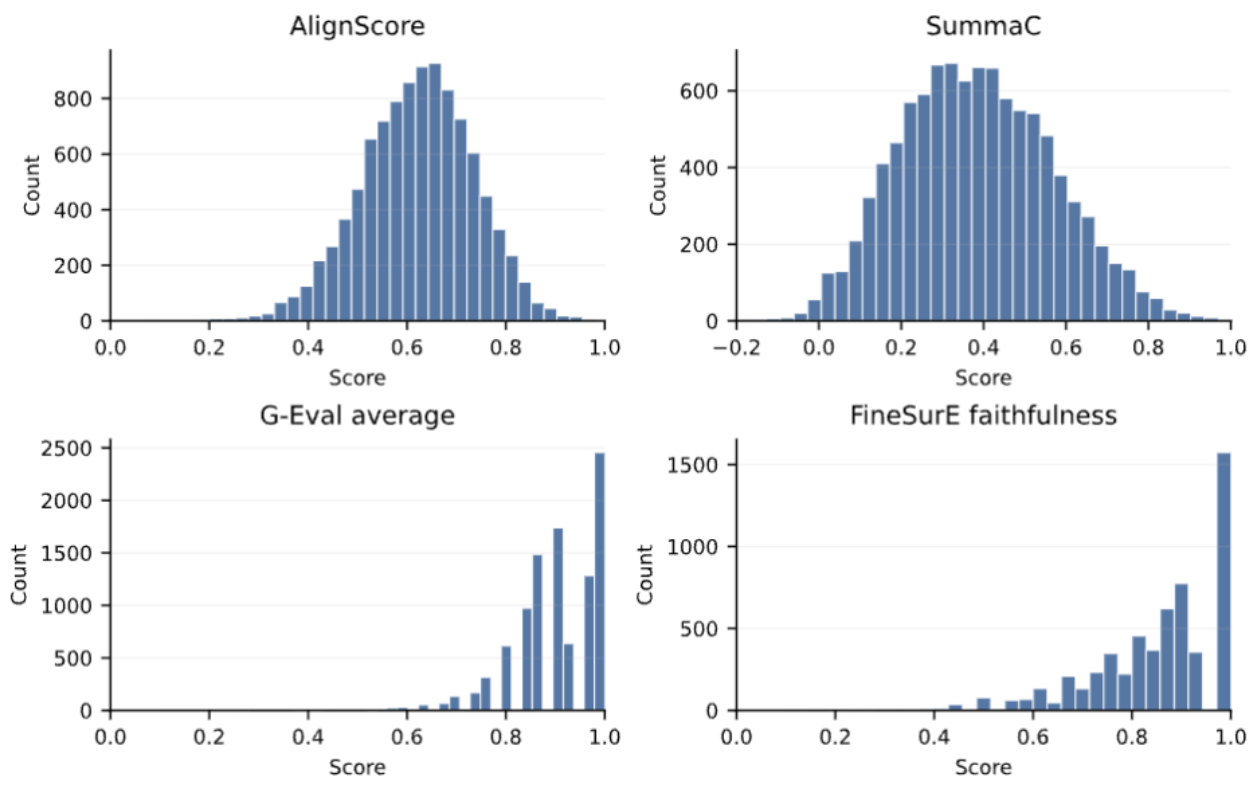}
\caption{Score distributions for author-written abstracts on the 10,000-example reference-quality analysis set.}
\label{fig:abstract-quality-distributions}
\end{figure}

Figure~\ref{fig:abstract-quality-distributions} and Table~\ref{tab:abstract-quality-results} summarize the reference-quality scores for the 10,000 author-written abstracts. The metrics use different scales, so the values should be interpreted within each metric rather than compared directly across metrics. Overall, many abstracts receive high scores, especially under G-Eval and FineSurE, but the distributions also show that reference quality varies across examples. 

\begin{table}[h]
\centering
\small
\begin{tabular}{lrcccc}
\hline
\textbf{Metric} & \textbf{Valid} & \textbf{Mean} & \textbf{P10} & \textbf{Med.} & \textbf{P90} \\
\hline
AlignScore & 10,000 & 0.622 & 0.473 & 0.626 & 0.765 \\
SummaC & 10,000 & 0.385 & 0.149 & 0.377 & 0.632 \\
G-Eval avg & 10,000 & 0.902 & 0.800 & 0.900 & 1.000 \\
FineSurE & 5,775 & 0.850 & 0.667 & 0.875 & 1.000 \\
\hline
\end{tabular}
\caption{Distribution of reference-quality scores on the 10,000-example abstract analysis set. G-Eval avg is the mean of coherence, factual consistency, and relevance. FineSurE is reported only over successful outputs.}
\label{tab:abstract-quality-results}
\end{table}

AlignScore has a median score of 0.626, while the bottom decile falls below 0.473. G-Eval also assigns high average scores, with a median of 0.900. FineSurE gives an average faithfulness score of 0.850 over successful outputs, with a bottom decile below 0.667. FineSurE succeeds on 5,775 examples, which suggests that it is harder to apply reliably at this scale and input length.

The metrics show only partial agreement, as shown in Table~\ref{tab:abstract-quality-correlations}. G-Eval average and FineSurE have the strongest association among the main metrics ($\rho=0.518$). AlignScore is positively but more weakly correlated with FineSurE ($\rho=0.293$) and G-Eval average ($\rho=0.231$). In contrast, SummaC has weak correlation with the other metrics, including AlignScore ($\rho=0.067$), G-Eval average ($\rho=-0.038$), and FineSurE ($\rho=0.025$).

These results suggest that author-written abstracts are useful training targets, but their reference quality varies across articles. Some abstracts are strongly aligned with their source articles, while others receive substantially lower scores. The limited agreement between metrics also indicates that no single metric should be treated as a complete measure of reference quality. This motivates the data-selection experiments in Section~\ref{sec:data_selection}, where we test whether these quality signals can be used to select better supervision examples.

\section{Quality-Aware Data Selection}
\label{sec:data_selection}
Next, we investigate whether reference-quality signals are useful for sampling training data. The experiments in this section compare metric-selected subsets with random or larger unfiltered subsets under the same model and evaluation settings.

\subsection{Model and Training Setup}
For all Supervised Fine-Tuning (SFT) experiments, we use Qwen2.5-3B as the base model. Each article body is inserted into a fixed summarization prompt:
\begin{quote}
\small
\texttt{Summarize the following document.\textbackslash n\textbackslash n\{text\}\textbackslash n\textbackslash nSummary:}
\end{quote}
The target output is the author-written abstract. All main SFT runs are trained for one epoch. More details regarding the experimental setup are presented in Appendix~\ref{app:experimental_details}.


\begin{table}[t]
\centering
\small
\begin{tabular}{lrrrr}
\hline
 & \textbf{Align} & \textbf{SummaC} & \textbf{G-Eval} & \textbf{FineSurE} \\
\hline
AlignScore & 1.000 & 0.067 & 0.231 & 0.293 \\
SummaC & 0.067 & 1.000 & -0.038 & 0.025 \\
G-Eval avg. & 0.231 & -0.038 & 1.000 & 0.518 \\
FineSurE & 0.293 & 0.025 & 0.518 & 1.000 \\
\hline
\end{tabular}
\caption{Spearman correlations between reference-quality metrics.}
\label{tab:abstract-quality-correlations}
\end{table}

\subsection{Evaluation Metrics}
\label{sec:eval-metrics}
All fine-tuned models and zero/few-shot baselines are evaluated on the same 500-example test set, sampled from PMC-Large and disjoint from the 10,000-example analysis set and training pools. We generate model summaries from the processed article body and evaluate them against the source document using the four metrics introduced in Section \ref{subsec:metrics}. FineSurE is reported over successfully parsed outputs, with the success rate reported alongside the score. For compact comparison within each table, we also report an {\em overall} score based on average rank. For each metric, models are ranked from best to worst, and the ranks are averaged across AlignScore, SummaC, G-Eval, and FineSurE. We report a linearly rescaled version of this average rank in the 0--1 range, where higher is better. Overall is only used for comparison within the same table.

\begin{table*}
\centering
\small
\begin{tabular}{lccccc}
\hline
\textbf{Model} & \textbf{Overall $\uparrow$} & \textbf{Align $\uparrow$} & \textbf{SummaC $\uparrow$} & \textbf{G-Eval $\uparrow$} & \textbf{FineSurE $\uparrow$ / Succ.} \\
\hline
Zero-shot & 0.444 & 0.735 & \textbf{0.462} & 0.822 & 0.847 / 76.6 \\
One-shot & 0.306 & 0.712 & 0.409 & 0.523 & 0.616 / 45.0 \\
Two-shot & 0.028 & 0.652 & 0.378 & 0.419 & 0.526 / 33.6 \\
\hline
Random 1K & 0.361 & 0.812 & 0.377 & 0.831 & 0.921 / 93.6 \\
All 5K & 0.583 & 0.825 & 0.382 & 0.845 & 0.921 / 95.8 \\
\hline
AlignScore 1K & 0.778 & \textbf{0.850} & 0.388 & 0.839 & \textbf{0.932} / 95.2 \\
FineSurE 1K & 0.667 & 0.834 & 0.395 & 0.834 & 0.928 / 94.0 \\
SummaC 1K & 0.472 & 0.819 & 0.403 & 0.819 & 0.908 / 94.2 \\
G-Eval 1K & 0.556 & 0.819 & 0.382 & \textbf{0.848} & 0.921 / 93.6 \\
Aggregate 1K & \textbf{0.806} & 0.844 & 0.398 & 0.845 & 0.931 / 94.8 \\
\hline
\end{tabular}
\caption{Experiment 1 results on the 500-example test set. Overall is computed as the average rank across AlignScore, SummaC, G-Eval, and FineSurE. FineSurE is reported over successful outputs, with the success rate shown after the slash.}
\label{tab:exp1-results}
\end{table*}

\subsection{Quality filter improves sample efficiency}
\label{exp1}
The first training experiment tests whether reference-quality metrics can identify better supervision samples. We use 5,000 randomly sampled examples from the 10,000-example reference-quality analysis set as a fixed candidate pool for training-data selection. Each candidate abstract is scored against its article body using AlignScore, FineSurE, SummaC-ZS PubMedBERT, and G-Eval.


For each metric, we rank the 5,000 candidate examples by score and select the top 1,000 examples as a training subset. 
We train an SFT model for each metric-selected subset. In addition, we also construct an aggregate score criterion by converting each metric score into a percentile rank within the candidate pool and averaging the four percentile ranks. We then select the 1,000 examples with the highest aggregate score and train a fifth SFT model on the resulting subset.

We compare these models with two additional SFT baselines, one trained on a randomly sampled subset of 1,000 examples and one trained on the full 5,000 example pool. This allows us to test whether metric-selected training sets outperform random training sets at the same size, and whether a smaller selected subset can approach or exceed a model trained on all available examples. We also evaluate the base language model with zero-shot, one-shot, and two-shot prompting\footnote{The one-shot and two-shot settings use larger maximum context lengths so that the demonstrations and test document fit in the input context.}.


Table~\ref{tab:exp1-results} shows the results of the first training experiment. Fine-tuning improves substantially on AlignScore, G-Eval, and FineSurE, but not on SummaC, where the zero-shot baseline obtains the highest score. SummaC is based on natural language inference (NLI), meaning that it evaluates whether summary statements can be inferred from the source document. One possible explanation is that zero-shot summaries make more general claims that are easier for the biomedical NLI model to verify, whereas fine-tuned models may generate more specific abstract-like statements that improve other quality dimensions but are harder to classify as directly entailed. This is also consistent with the weak correlation between SummaC and the other metrics in Table~\ref{tab:abstract-quality-correlations}. Zero-shot prompting is the strongest prompting baseline, while one-shot and two-shot prompting perform worse, suggesting that adding demonstrations does not help in this long-document setting.

Among SFT models, the randomly selected 1K subset already improves over the prompting baselines on AlignScore, G-Eval, and FineSurE. However, selecting examples by reference quality further improves performance. The aggregate 1K model outperforms the random 1K model on all four metrics and also matches or exceeds the full 5K model on most metrics.

The individual selection metrics show different behavior. The AlignScore selected model obtains the highest AlignScore and FineSurE scores, while the G-Eval selected model obtains the highest G-Eval score. The SummaC selected model obtains the highest SummaC score among SFT models, but it still remains below the zero-shot baseline on SummaC, suggesting that SummaC-based selection is less effective in this setting. Overall, the aggregate selection gives the strongest balance across metrics, supporting the use of multiple quality signals rather than relying on a single metric alone.

\begin{table*}
\centering
\small
\begin{tabular}{lccccc}
\hline
\textbf{Model} & \textbf{Overall $\uparrow$} & \textbf{Align $\uparrow$} & \textbf{SummaC $\uparrow$} & \textbf{G-Eval $\uparrow$} & \textbf{FineSurE $\uparrow$ / Succ.} \\
\hline
\multicolumn{6}{l}{\textit{AlignScore selection from 100K}} \\
Random 1K & 0.036 & 0.810 & 0.388 & 0.838 & 0.906 / 94.8 \\
AlignScore 1K & 0.625 & \underline{\textbf{0.887}} & \textbf{0.405} & 0.824 & 0.949 / 95.8 \\
Random 2K & 0.125 & 0.816 & 0.384 & 0.846 & 0.916 / 95.0 \\
AlignScore 2K & 0.571 & 0.879 & 0.395 & 0.841 & \textbf{0.957} / 95.0 \\
Random 5K & 0.179 & 0.825 & 0.392 & 0.843 & 0.913 / 93.8 \\
AlignScore 5K & \textbf{0.679} & 0.883 & 0.398 & 0.849 & \textbf{0.957} / 95.4 \\
Random 10K & 0.250 & 0.833 & 0.390 & 0.848 & 0.917 / 94.2 \\
AlignScore 10K & 0.589 & 0.866 & 0.397 & 0.868 & 0.947 / 94.4 \\
All 100K & 0.429 & 0.828 & 0.391 & \underline{\textbf{0.874}} & 0.935 / 95.2 \\
\hline
\multicolumn{6}{l}{\textit{Aggregate selection inside the top 10K AlignScore pool}} \\
Agg-All 1K & 0.536 & 0.867 & 0.400 & 0.852 & 0.947 / 94.2 \\
Agg-NoSummaC 1K & 0.429 & 0.864 & 0.392 & 0.856 & 0.945 / 94.4 \\
Agg-All 2K & 0.679 & 0.877 & 0.401 & 0.855 & 0.954 / 94.6 \\
Agg-NoSummaC 2K & 0.786 & \textbf{0.879} & 0.401 & 0.864 & 0.957 / 96.2 \\
Agg-All 5K & \underline{\textbf{0.839}} & 0.876 & \underline{\textbf{0.406}} & 0.861 & \underline{\textbf{0.961}} / 93.6 \\
Agg-NoSummaC 5K & 0.750 & 0.876 & 0.399 & \textbf{0.867} & 0.958 / 95.2 \\
\hline
\end{tabular}
\caption{Results for the large candidate pool experiments on the 500-example test set. Overall is computed within this table using the average-rank score described in Section~\ref{sec:eval-metrics}. Bold marks the best score within each experimental block, and underlining marks the best score across the full table.}
\label{tab:large-selection-results}
\end{table*}

\subsection{Quality selection remains effective at scale}
\label{exp2}
We next examine whether quality-based selection remains useful when the candidate pool is scaled from 5,000 to 100,000 examples. Computing all metrics over 100,000 long articles is expensive, especially for LLM-based metrics such as G-Eval and FineSurE. We therefore use AlignScore as a selection metric. AlignScore can be run locally and was competitive with the strongest selection methods in Section~\ref{exp1}, making it a practical choice for this purpose.

We sample a 100,000 example candidate pool and compute AlignScore for each abstract against its corresponding article body. We then construct selected training subsets by taking the top 1,000, 2,000, 5,000, and 10,000 examples according to AlignScore. For each training size, we also construct a random subset of the same size from the same candidate pool. In addition to the selected and random subsets, we train a model on the full 100,000 example pool.

The upper block of Table~\ref{tab:large-selection-results} compares AlignScore selected subsets against random subsets from the same 100,000 example pool. At every matched training size, the AlignScore selected model obtains a higher overall score than the corresponding random baseline. Models trained on quality-selected subsets also consistently improve AlignScore and FineSurE, showing that the benefit of quality-based selection remains visible when the candidate pool is scaled up.

The quality-selected subsets also compare favorably with training on the full 100K pool. The 5K AlignScore selected model is the strongest model in the first block by Overall score, and it outperforms the 100K model on AlignScore, SummaC, and FineSurE. The 100K model obtains the highest G-Eval score, which suggests that larger unfiltered training data may still improve some aspects of generation quality. Overall, these results show that more data is not always better than better selected data.

\subsection{Two-stage quality-based selection improves performance}
Finally, we test whether the large-scale setting can benefit from the multi-metric selection strategy that performed best in Section~\ref{exp1}. The goal is to test whether a more expensive second-stage selection step can further improve the quality of the training data. We keep AlignScore as a cheap and scalable first-stage filter and test whether reranking its highest-scoring examples with additional metrics further improves training-data selection. 


We begin with the top 10,000 examples selected by AlignScore from the 100,000 example pool and we compute the remaining metrics for these 10,000 examples. 
We consider two aggregate variants: \textsc{Aggregate-All} which combines AlignScore, FineSurE, SummaC, and G-Eval and \textsc{Aggregate-NoSummaC} which combines AlignScore, FineSurE, and G-Eval, excluding SummaC, since earlier results showed that it behaved less consistently than the other metrics. For each aggregate variant, we select the top 1K, 2K, and 5K examples and train separate SFT models.

Results are shown in the lower block of Table~\ref{tab:large-selection-results}. \textsc{Aggregate-All} at 5K examples obtains the best Overall, SummaC and FineSurE scores across the full table, as shown by the underlined entries.  Removing SummaC improves G-Eval at each matched size, but it is not uniformly better across the other metrics. Overall, these results support a two-step selection strategy: first use AlignScore as a cheap and scalable filter at scale, and then use multiple metrics to refine the selected training subset.

\section{Conclusions}
\label{sec:conclusions}

In this work, we studied reference quality in long-document scientific summarization. We introduced PMC-Large, a large-scale PMC article-abstract dataset that preserves document structure for long-context language models, and used it to examine whether author-written abstracts can be treated as uniformly reliable references. Our analysis shows that author-written abstracts vary in their alignment with the full article, and that automatic metrics only partially agree on reference quality.

We also showed that these quality signals can improve training data selection. Metric-selected subsets consistently outperform random subsets at matched sizes, and smaller selected subsets can match or exceed much larger unfiltered training sets. Overall, our results suggest that author -written abstracts are useful but not uniformly reliable supervision, and that quality-aware selection is a simple way to improve training efficiency and factuality-oriented performance.

\section*{Limitations}
\label{sec:limitations}
While our results show that reference-quality signals are useful for scientific summarization, several aspects of the problem remain outside the scope of this work. We rely on automatic metrics, which may not capture all aspects of summary quality and were not specifically designed for biomedical and life science articles, so human evaluation would provide stronger validation. Due to computational cost and resource constraints, we compute quality scores and run fine-tuning experiments only on subsets of PMC-Large rather than on the full 1.88 million article--abstract pairs. Our experiments also use one base model and one main training setup, so the results may not fully generalize to larger models, other architectures, or different fine-tuning settings. In addition, the dataset is limited to English PMC Open Access articles with specific licensing and length filters, which may affect generalization to other domains, languages, or summary types. Finally, selecting examples with automatic metrics may favor the behavior of those metrics and reduce diversity in the training data.

\section*{Ethical Considerations}
\label{sec:ethical_considerations}
Our work studies the reliability of author-written abstracts and the use of automatic quality signals for training-data selection. Although these methods may improve factual consistency in scientific summarization, the resulting systems can still generate unsupported, incomplete, or misleading summaries. Generated summaries should therefore not be treated as authoritative scientific or clinical evidence without expert verification.

The dataset may also reflect existing biases in the biomedical literature. Since our training targets are author-written abstracts, any inaccuracies or interpretive bias present in the original abstracts may also be propagated to summarization systems trained on this data.

\section{Acknowledgements}
We used ChatGPT to help revise and improve the wording of the manuscript and to support the development of some parts of the code. The authors reviewed, verified and edited all AI-assisted outputs and take full responsibility for the final manuscript and implementation.

\bibliography{custom}

\appendix

\section{Additional Dataset Statistics}
\label{app:dataset_statistics}

\begin{figure}[H]
\centering
\includegraphics[width=0.48\linewidth]{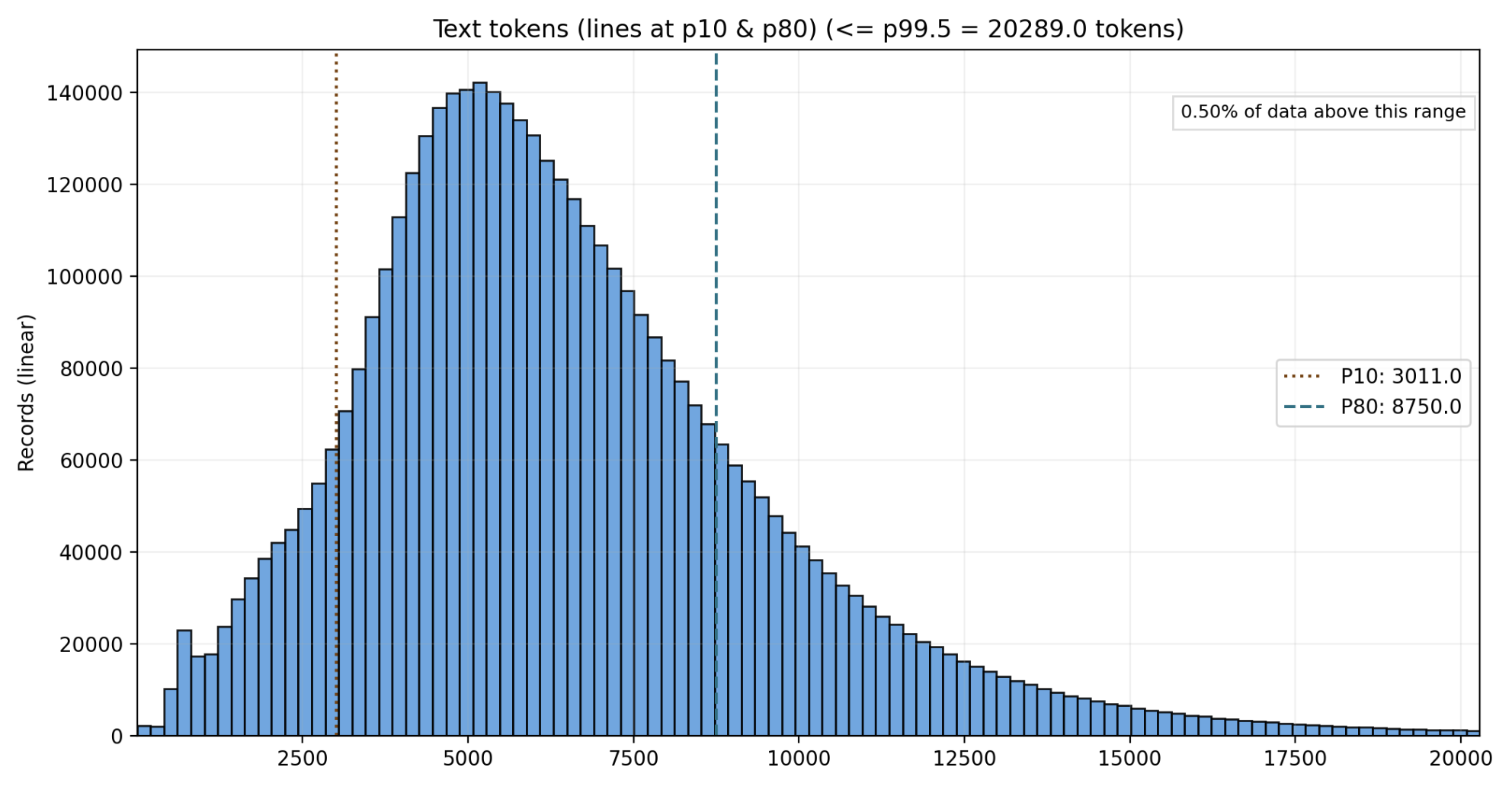} \hfill
\includegraphics[width=0.48\linewidth]{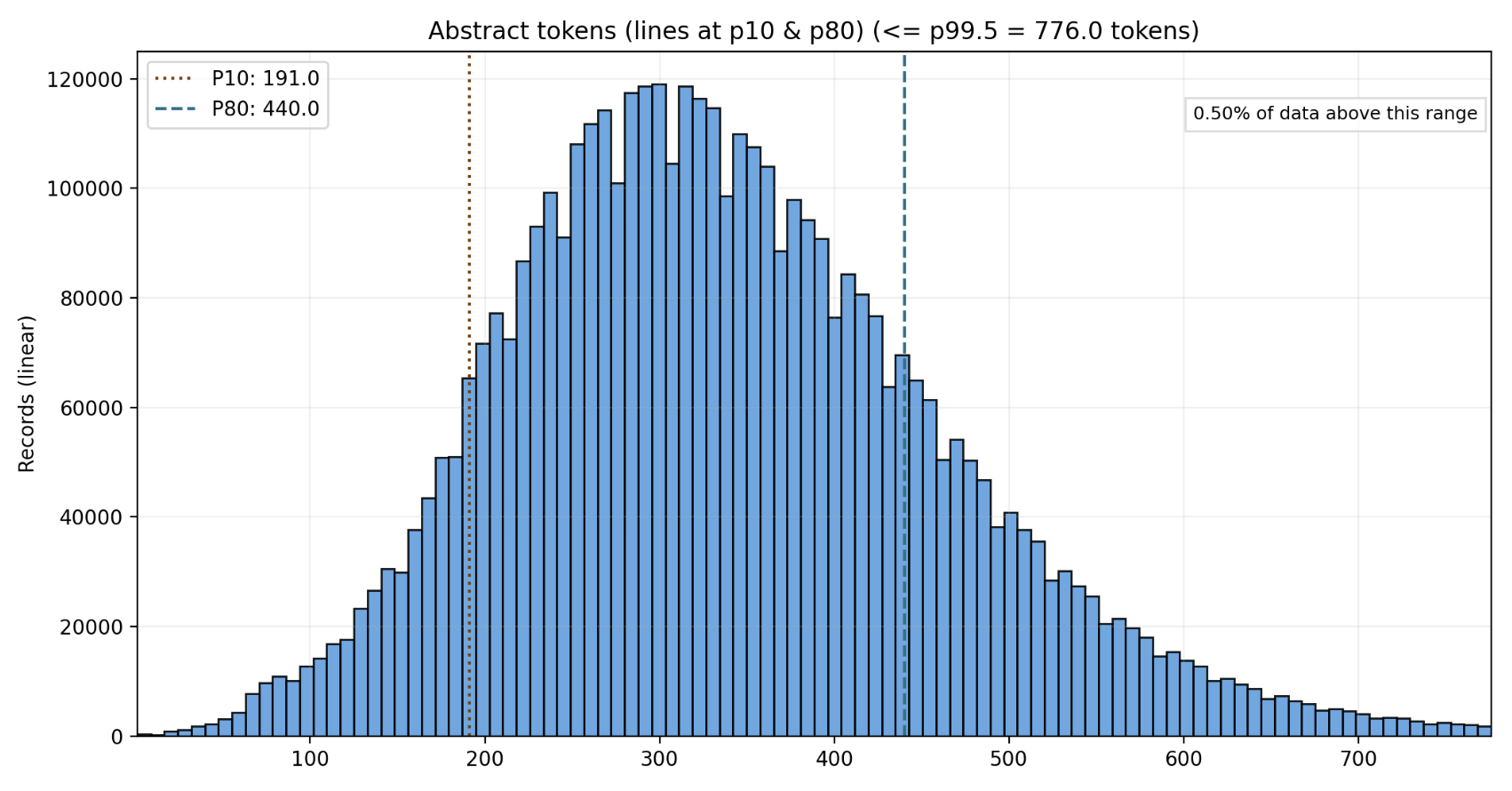}
\caption{Token-count distributions for article bodies and abstracts. These distributions are used to define the 10th-80th percentile length range retained in the final dataset.}  \label{fig:token-count}
\end{figure}

This section provides additional corpus statistics supporting the dataset construction described in Section~\ref{sec:dataset}. 
Figure~\ref{fig:token-count} shows the token-count distributions for article bodies and abstracts. 
These distributions are used to define the 10th-80th percentile length range retained in the final dataset. 

\begin{table*}
\centering
\small
\begin{tabular}{p{2cm} p{3cm} p{4cm} p{3cm}}
\hline
PMC ID & Original PMC & PubMed~\citep{cohan-etal-2018-discourse} & PMC-Large (Ours) \\
\hline
PMC4165415 &
having $\geq$ 3 &
having 3 &
having $\geq$ 3 \\


PMC3881782 &
solution at 37ºC &
solution at 37c &
solution at 37ºC \\

PMC4540752 &
1–2,3–7,8+ &
12 , 37 , 8 + &
1–2,3–7,8+ \\

PMC4540752 &
$W_{ab}=S_{ab}/K_a$ &
wab = sab / ka &
$W_{ab}=S_{ab}/K_a$ \\


PMC416487 &
(1–2.5\%) \underline{[3,4,32,33]} &
( 12.5\%) . &
(1–2.5\%) [CIT\_REF] \\

PMC5008198 &
of nano-TiO$_2$ &
of nanotio2 &
of nano-TiO$_2$ \\

\hline
\end{tabular}
\caption{Examples of text normalization. Our preprocessing preserves scientific symbols where possible and normalizes citations, figure/table mentions, and mathematical notation into a consistent text representation.}
\label{tab:normalization_examples}
\end{table*}

\section{Additional Experimental Details}
\label{app:experimental_details}

This section reports implementation details for the supervised fine-tuning experiments described in Section~\ref{sec:data_selection}. 
In an initial 3-epoch pilot, validation loss improved during approximately the first epoch and then increased while training loss continued to decrease, suggesting overfitting. We therefore train all models for one epoch in the main experiments. During training, we evaluate on 500 validation examples and select the best checkpoint according to validation loss. We train on one NVIDIA H100 80GB GPU using bf16 precision and gradient checkpointing. For generation, we use greedy decoding with a maximum of 768 new tokens. Table~\ref{tab:training-hparams} lists the main training, generation, and hardware settings used in the experiments. 

\begin{table}[H]
\centering
\small
\begin{tabular}{ll}
\hline
\textbf{Hyperparameter} & \textbf{Value} \\
\hline
Base model & \texttt{\texttt{Qwen2.5-3B}} \\
Epochs & 1 \\
Learning rate & $1.32 \times 10^{-5}$ \\
Batch size & 1 \\
Gradient accumulation & 1 \\
Weight decay & 0.0374 \\
Warmup steps & 163 \\
Max gradient norm & 0.8873 \\
Evaluation / save interval & 100 steps \\
Max sequence length & 12,000 \\
Max new tokens & 768 \\
Precision & bf16 \\
Hardware & NVIDIA H100 80GB \\
Seed & 42 \\
\hline
\end{tabular}
\caption{Main supervised fine-tuning hyperparameters.}
\label{tab:training-hparams}
\end{table}

\section{Preprocessing Examples}
\label{app:preprocessing_examples}

This section provides qualitative examples supporting the preprocessing decisions described in Section~\ref{sec:dataset}. 
The examples compare our preprocessing with the released PubMed dataset and illustrate two main differences: content-aware section filtering and text normalization. 
They are intended as qualitative examples of preprocessing behavior rather than as a separate quantitative evaluation.

Table~\ref{tab:section_filtering_examples} shows examples of non-narrative sections that are retained in the released PubMed dataset but excluded by our content-aware section filter. 
Table~\ref{tab:normalization_examples} provides examples of normalization decisions, including preserving scientific notation where possible and replacing inline references with standardized placeholders.

\begin{table}
\centering
\small
\begin{tabular}{p{1.7cm}p{6.8cm}}
\hline
PMC ID & Non-narrative sections retained in PubMed \\
\hline
PMC4165415 &
Funding; Footnotes \\
PMC165589 &
List of Abbreviations; Competing Interests; Authors' Contributions; Pre-publication history; Supplementary Material; Acknowledgements \\
PMC4540752 &
Electronic supplementary material; Additional files; Acknowledgements; Compliance with ethical guidelines \\
PMC4240245 &
Data availability; Consent \\
PMC4262794 &
Ethical Considerations; Conflict of Interest; Funding; Authorship; Research Ethics Statement \\
\hline
\end{tabular}%

\caption{Examples of non-narrative sections that remain in the released PubMed dataset but are removed by our content-aware section filter. We report only the differing sections rather than the complete section lists.}
\label{tab:section_filtering_examples}
\end{table}

\label{sec:appendix}

\end{document}